\documentclass[sigconf,10pt]{acmart}
\usepackage{algorithm, algpseudocode}

\newcommand*{\thead}[1]{\multicolumn{1}{c|}{\bfseries #1}}
\begin{document}

\title{A Real-time Edge-AI System for Reef Surveys}

\settopmatter{printacmref=false} 
\renewcommand\footnotetextcopyrightpermission[1]{} 
\pagestyle{plain} 
\setcopyright{none}





\iftrue
\author{Yang Li$^\dag$ \quad Jiajun Liu$^\dag$\quad Brano Kusy$^\dag$\quad Ross Marchant$^{\circ\dag}$\quad  Brendan Do$^\dag$\quad \newline Torsten Merz$^\dag$ Joey Crosswell$^\star$\quad  Andy Steven$^\star$\quad Lachlan Tychsen-Smith$^\dag$ \newline \quad David Ahmedt-Aristizabal$^\dag$ \quad Jeremy Oorloff$^\dag$ \quad Peyman Moghadam$^\dag$ \newline Russ Babcock$^\star$\quad Megha Malpani$^\diamond$ \quad Ard Oerlemans$^\diamond$}

\authornote{Affiliations: $^\dag$CSIRO's Data61, $^\star$CSIRO Oceans \& Atmosphere, $^\circ$Queensland University of Technology, $^\diamond$Google \newline Emails: \{yang.li1,\ jiajun.liu,\ brano.kusy,\ ross.marchant,\ brendan.do,\ torsten.merz,\ joey.crosswell,\ andy.steven,\ laclan.tychsen-smith,\  david.ahmedtaristizabal,\ jeremy.oorloff, \ peyman.moghadam, \ russ.babcock\}@csiro.au,\ \{mmalpani,\ ardoerlemans\}@google.com}
\fi

\begin{abstract}
  Crown-of-Thorn Starfish (COTS) outbreaks are a major cause of coral loss on the Great Barrier Reef (GBR) and substantial surveillance and control programs are ongoing to manage COTS populations to ecologically sustainable levels. In this paper, we present a comprehensive real-time machine learning-based underwater data collection and curation system on edge devices for COTS monitoring. In particular, we leverage the power of deep learning-based object detection techniques, and propose a resource-efficient COTS detector that performs detection inferences on the edge device to assist marine experts with COTS identification during the data collection phase. The preliminary results show that several strategies for improving computational efficiency (\textit{e.g.}, batch-wise processing, frame skipping, model input size) can be combined to run the proposed detection model on edge hardware with low resource consumption and low information loss.
\end{abstract}



\maketitle

\section{Introduction}
\label{sec:intro}

\begin{figure}
    \centering
    \includegraphics[width=0.9\columnwidth]{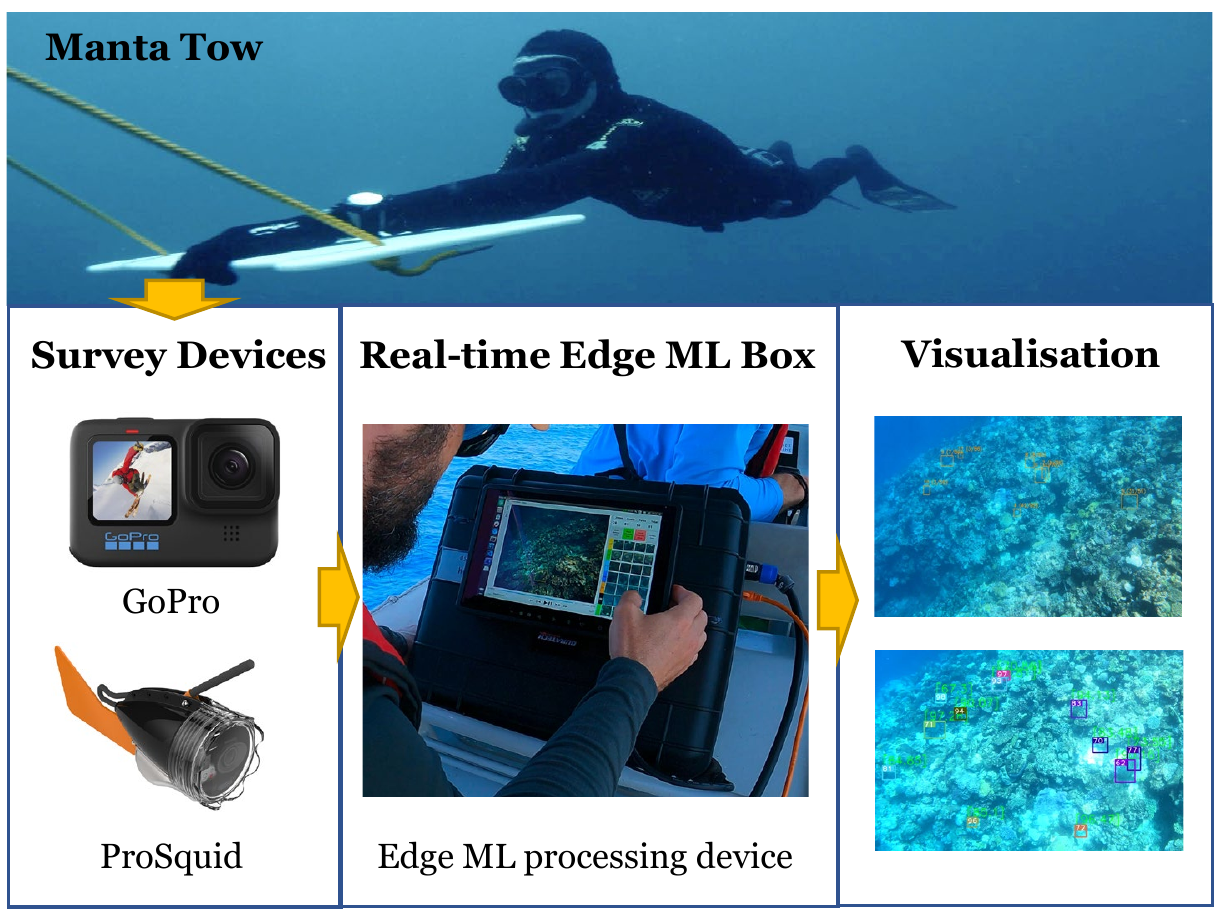}
    \caption{The edge ML box deployed on the boat facilitates reef surveys.}
    \vspace{-2em}
    \label{fig:edge_ml_box}
\end{figure}

\begin{figure*}[hbt!]
  \includegraphics[width=0.9\textwidth]{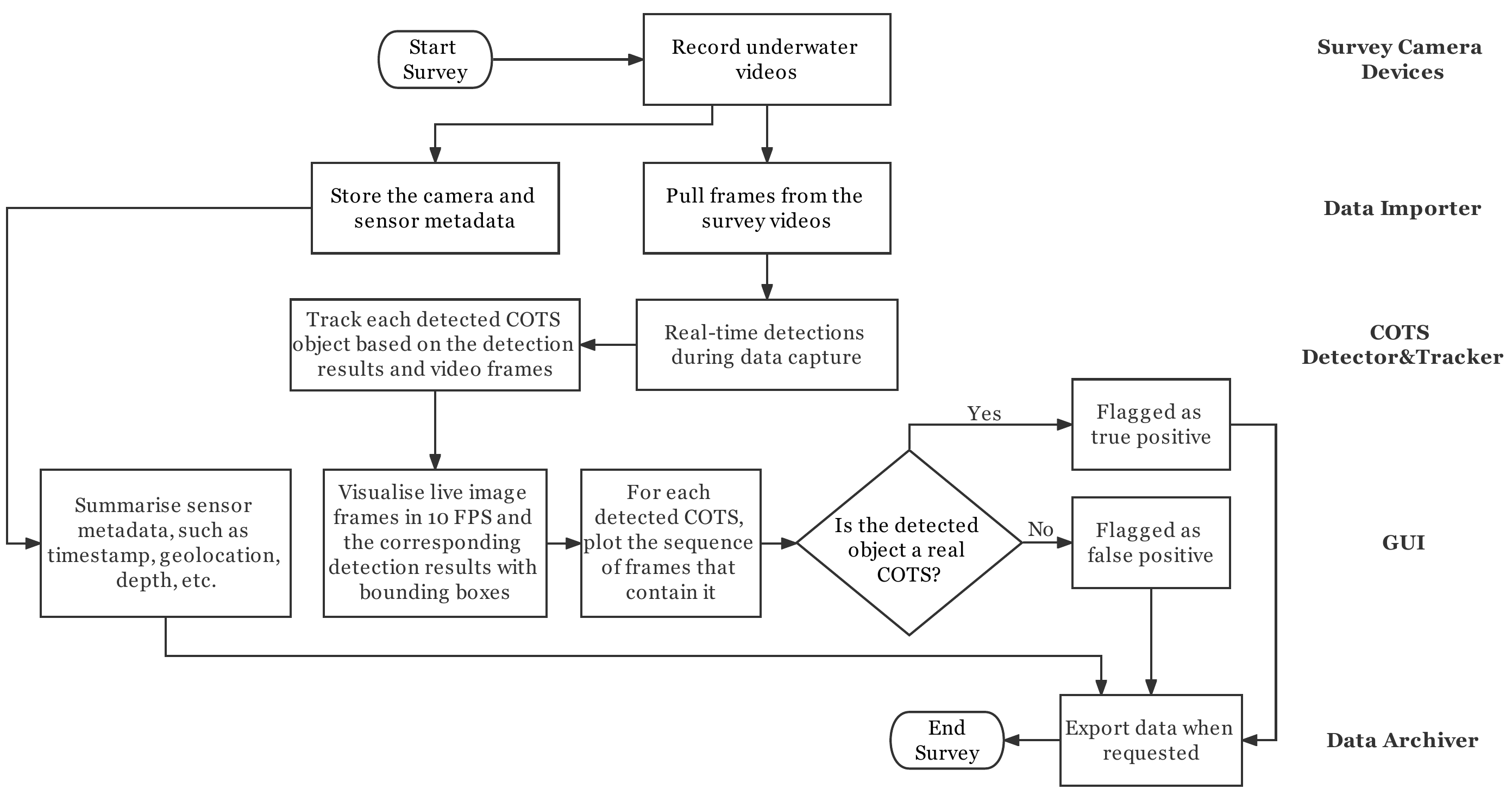}
    \vspace{-6pt}
  \caption{The overall workflow of data collection, curation, and analytics on our edge device.}
  \vspace{-1em}
  \label{fig:MLBox_Workflow}
\end{figure*}

Australia's Great Barrier Reef (GBR), a national icon and a World Heritage Site in Australia, has now been under threat because of the Crown-of-Thorns Starfish (COTS) population explosion. COTS are a common reef species that feed on corals and have been an integral part of coral reef ecosystems. However, their populations can explode to thousands of starfish on individual reefs, eating up most of the coral and leaving behind a white-scarred reef that will take years to recover. To protect the reef environment, Australian agencies are running a major monitoring and management program to track and control COTS populations to ecologically sustainable levels. 
One of the key survey methods used for COTS surveillance is the \textit{Manta Tow} method, in which a snorkel-diver is towed behind a boat to perform a visual assessment of the underwater habitat. Information from such surveys is used to identify early or existing outbreaks of COTS and provides key information for the decision support systems to better target the deployment of COTS control teams. Nevertheless, COTS are cryptic animals and like to hide, thus in many cases, only part of the starfish body might be visible. To help marine experts find COTS more effectively, we propose to collect underwater video data along the defined transect paths for COTS monitoring and deploy a machine learning-driven solution to find all visible COTS. 
Specifically, we introduce the \textbf{COTS} \textbf{U}nderwater \textbf{D}ata \textbf{A}nalytics \textbf{S}ystem (COTS-UDAS) which deploys a lightweight object detection model to automatically find COTS in the captured underwater video stream. In reef surveys, as shown in Figure \ref{fig:edge_ml_box}, we deploy our system onto the edge ML box that is built around an Nvidia Jetson Xavier device equipped with a GPS receiver. The entire box is powered by a portable battery for 4-hour operations on the boat in the ocean. The system allows marine experts to review the detection results in the field to improve survey speed and efficiency.

One challenge we faced in designing the COTS-UDAS was that most off-the-shelf real-time object detectors on server standard operations, such as 
EfficientDet~\cite{tan2020efficientdet}, FCOS~\cite{tian2019fcos} and YOLO \cite{cvpr21_scaled_yolo}, are compute-intensive. For example, YOLO takes about 20 billion floating point operations per second (FLOPS). It is thus challenging to run these object detectors on an edge device in a real-time manner without skipping a large number of frames. 
To this end, we discuss the factors that may affect the inference speed of the detector and propose a real-time target detection algorithm based on Scaled-YOLOv4~\cite{cvpr21_scaled_yolo} on edge devices. 
We further integrate an object tracker into the detection module, which infers the position of each COTS based on the consecutive frames by optical flow, and can be helpful in capturing those COTS that are only visible at specific camera angles.


\vspace{-6pt}
\section{COTS Underwater Data Analytics System}
We collected the data by adapting underwater survey cameras to the \textit{Manta Tow} method. Specifically, the survey camera is attached to the bottom of the manta tow board that the diver holds during the survey. Once the survey is started, the system follows the workflow demonstrated in Figure \ref{fig:MLBox_Workflow} to store the collected data and provide domain experts with real-time survey insights. In general, there are four main components in this workflow:\\
\textbf{Survey Recording Devices}: GoPro or ProSquid underwater recording cameras.\\
\textbf{Data Importer}: A launched process pulls frames from the survey video, converts them into JPG images, and writes out images including both image data and sensor metadata (\textit{i.e.}, timestamp and geolocation).\\
\textbf{COTS Detector\&Tracker}: Both COTS detector and tracker programs are pre-loaded when the survey starts. The detector consumes the new image frames coming out from the recording devices, and performs inferences. Then, the tracker adjusts each detected result based on the optical flow estimated from consecutive frames.\\
\textbf{Graphical User Interface (GUI)}: A desktop application installed on the edge device, which visualises live underwater data and detection results on the screen. The same detected COTS objects across the video frames are summarised in an image sequence. The marine experts can review those COTS image sequences during the survey, and select whether the detected results are true positive or false positive objects.\\
\textbf{Data Archiver}: A program that exports the curated data to a specified data drive.

\vspace{-7pt}
\section{COTS Detection and Tracking}
We developed our COTS detection model based on Scaled-YOLOv4 architecture \cite{cvpr21_scaled_yolo}. 
COTS are usually small and like to hide in corals, which makes them hard to identify. Therefore, in addition to the original training data, we applied a range of image augmentation techniques including flipping, scaling, rotation, and multi-image mix-up to create more data samples for the model training.

As mentioned in Section \ref{sec:intro}, direct adoption of deep detectors on edge devices will result in extremely slow inference speed. Thus, to minimise the detector's prediction latency, we propose to optimise the detector using the following strategies\footnote{Our inference pipeline is available at \url{https://github.com/tensorflow/models/tree/master/official/projects/cots_detector}.}:\\
\textbf{Input size}: Intuitively, large-sized images fed into the model increase the processing time, since the convolutional neural network (CNN) kernel needs more steps to scan through an image. Thus, reducing the size of the images obtained from the survey recording devices can significantly decrease the processing time.\\
\textbf{Batch Size}: Instead of processing images one by one in a video sequence, processing image frames in a batch-wise manner, \textit{i.e.}, a number of images are fed into the detector at the same time, can improve the computational efficiency.\\
\textbf{Frame Skipping}: Since COTS are moving very slowly, we observe that the same detected COTS objects appear in hundreds of frames in a video frame sequence, and their positions only change a little between two consecutive frames. Hence, selectively running the detector on video frames enhances the resource consumption efficiency on edge devices.\\
\textbf{Buffer Queue}: The different processing rates of the data importer, detector and tracker hinder the compute efficiency since some components have to wait until others finish. To alleviate this issue and enable parallel computing in our system, we introduce two buffer queues that store the input image frames and prediction results, respectively. In this way, the detector and tracker can run in parallel as long as the buffer queues are not empty.\\

Finally, we describe the details of how the strategies above are employed in our detection module in the following:\\
\textbf{COTS Object Detection}: To optimise the detector inference speed, the model performs detection of the queued images in batches with mixed precision. All the output detections that have above-threshold confidence scores are then stored in another queue shared with the COTS object tracker.\\
\textbf{COTS Object Tracking}: The COTS object tracker is running in a separate thread, which works in parallel with the detector. It obtains the detections from the shared queue chronologically and estimates each existing track's new object position in the current frame through the optical flow. For each detection, if it matches any existing track, it will be linked to that track, otherwise, it will be saved in a new track. The tracking results are finally returned to the system interface for visualisation and further analysis.

\section{Experiments}
\begin{table}[t]
\caption{Comparison of the prediction accuracy and latency of our detection model with different system parameters.}
\vspace{-6pt}
\label{tab:model_performance}
\resizebox{\columnwidth}{!}{
    \begin{tabular}{l|c|c}
    \thead{Model} & \textbf{F2 Score} & \textbf{FPS} \\
    \hline
    YOLOv4 (batch size 1, 1080p, w/o TensorRT) & 0.56     & 2 \\
    YOLOv4 (batch size 4, 1080p, w/o TensorRT) & 0.56     & 5  \\
    YOLOv4 (batch size 4, 720p, w/o TensorRT)  & 0.53     & 14  \\
    YOLOv4 (batch size 4, 720p, with TensorRT) & 0.52     & 22 
    \end{tabular}
}
\vspace{-1.5em}
\end{table}

To verify the efficiency and effectiveness of our detection model in COTS-UDAS, we conducted preliminary experiments on our public COTS dataset \cite{arxiv_cots_dataset}. In particular, it has three survey transect videos (24,554 images) in the training set and one video (11,005 images) in the test set. We first train the detector on a cloud server with Nvidia RTX 3090 GPU and then deploy it on Xavier during the test phase. The evaluation metrics F2 score \cite{f2_score} and FPS are adapted for the prediction effectiveness and efficiency, respectively. We report the model performance over various settings in Table \ref{tab:model_performance}. From the table, we can have the following observations:\\
(1) Our model, with a higher input size, \textit{i.e.}, 1080 $\times$ 1080, achieves the best performance as more detailed patterns of COTS can be captured by the model. However, the large model size results in high processing latency (2FPS) on edge devices. Even when the batch processing strategy is employed, the processing speed (\textit{i.e.}, 5 FPS when batch size is 4) still cannot meet the real-time requirements.\\
(2) By scaling down the input size from 1080p to 720p, the model's F2 score performance drops slightly from 0.56 to 0.53 but significantly reduces the prediction latency per frame, which increases from 5 FPS to 14 FPS. \\
(3) We further optimise the computational graph generated in TensorFlow with TensorRT. TensorRT brings a 57\% inference speed boost for the 720p model (\textit{i.e.}, from 14 FPS to 22 FPS) and still preserves decent prediction accuracy (\textit{i.e.}, 0.52 F2 score) in the meantime.

\section{Conclusion}
In this paper, we present a real-time underwater analytics system that employs object detection techniques to facilitate the data collection and curation process for COTS monitoring. From the preliminary experimental results, we verify the efficacy of the designed strategies for real-time processing. In the future, we plan to explore more resource-efficient techniques and extend our system to other reef survey activities such as seagrass coverage estimation and multi-class marine species detection.

\bibliographystyle{ACM-Reference-Format}
\bibliography{main}


\end{document}